\documentclass[letterpaper, 10 pt, conference]{ieeeconf}  % Comment this line out if you need a4paper

\IEEEoverridecommandlockouts                              % This command is only needed if 
\usepackage{graphicx} % for pdf, bitmapped graphics files
\usepackage{amsmath} % assumes amsmath package installed
\usepackage{amssymb}  % assumes amsmath package installed
\usepackage{xcolor}

\title{\LARGE \bf
Efficient Decentralized Visual Place Recognition \\ From Full-Image Descriptors}

\author{Titus Cieslewski and Davide Scaramuzza% <-this % stops a space
\thanks{This is a self-published paper that accompanies our original work \cite{cieslewski2017efficient} as well as the ICRA 2017 Workshop on Multi-robot Perception-Driven Control and Planning \cite{icra2017multirobotws}}% <-this % stops a space
}

\begin{document}

\maketitle
\thispagestyle{empty}
\pagestyle{empty}

%%%%%%%%%%%%%%%%%%%%%%%%%%%%%%%%%%%%%%%%%%%%%%%%%%%%%%%%%%%%%%%%%%%%%%%%%%%%%%%%
\begin{abstract}

In this paper, we discuss the adaptation of our decentralized place recognition method described in \cite{cieslewski2017efficient} to full-image descriptors.
As we had shown, the key to making a scalable decentralized visual place recognition lies in exploting deterministic key assignment in a distributed key-value map.
Through this, it is possible to reduce bandwidth by up to a factor of $n$, the robot count, by casting visual place recognition to a key-value lookup problem.
In \cite{cieslewski2017efficient}, we exploited this for the bag-of-words method \cite{sivic2003video, nister2006scalable}.
Our method of casting bag-of-words, however, results in a complex decentralized system, which has inherently worse recall than its centralized counterpart.
In this paper, we instead start from the recent full-image description method NetVLAD \cite{arandjelovic2016netvlad}.
As we show, casting this to a key-value lookup problem can be achieved with k-means clustering, and results in a much simpler system than \cite{cieslewski2017efficient}.
The resulting system still has some flaws, albeit of a completely different nature:
it suffers when the environment seen during deployment lies in a different distribution in feature space than the environment seen during training.

\end{abstract}

%%%%%%%%%%%%%%%%%%%%%%%%%%%%%%%%%%%%%%%%%%%%%%%%%%%%%%%%%%%%%%%%%%%%%%%%%%%%%%%%
\section{Introduction}

In \cite{cieslewski2017efficient}, we have shown how the data exchange incurred in decentralized visual place recognition can be reduced by a factor of up to $n$, the robot count.
This can be achieved by casting the place recognition problem to a key-value lookup problem, which can be efficiently distributed using deterministic key-to-peer assignment, as is for example common in distributed hash tables \cite{stoica2001chord, rowstron2001pastry}.
In \cite{cieslewski2017efficient}, we have thus cast the bag-of-words (BoW) place recognition method \cite{sivic2003video, nister2006scalable} used in \cite{GalvezTRO12, mur2015orb}.
In broad strokes, this is how the resulting method works:
\begin{enumerate}
\item Before deployment, deterministically assign words of the visual vocabulary to the different robots.
\item When querying place recognition of an image frame, calculate the BoW vector and split it up into partial BoW vectors such that one partial BoW vector can be sent to each robot $r$, containing the coefficients of the words assigend to $r$.
\item The robots receive and process each their own partial query, returning the identity of the single frame which best matches the query frame according to the partial BoW vector.
They also store the query, making it available as a result for subsequent queries.
\item Gather all partial results and determine which frame is most consistently returned as result.
\item Send a full query to the robot that has observed that frame for geometric verification.
\end{enumerate}
We have shown that this methods results in a bandwidth reduction of up to $n$ (depending on the network infrastructure), while reducing recall by $10-20\%$ depending on the robot count.
A lot of the recall reduction is due to steps 3) and 4) of the method, which are based on a simplifying assumption that we do not yet fully understand.
See Sec. IV C. and Fig. 4 of \cite{cieslewski2017efficient} for a detialed discussion.

\section{Methodology}

\begin{figure}
  \centering
  \includegraphics[width=.5\columnwidth]{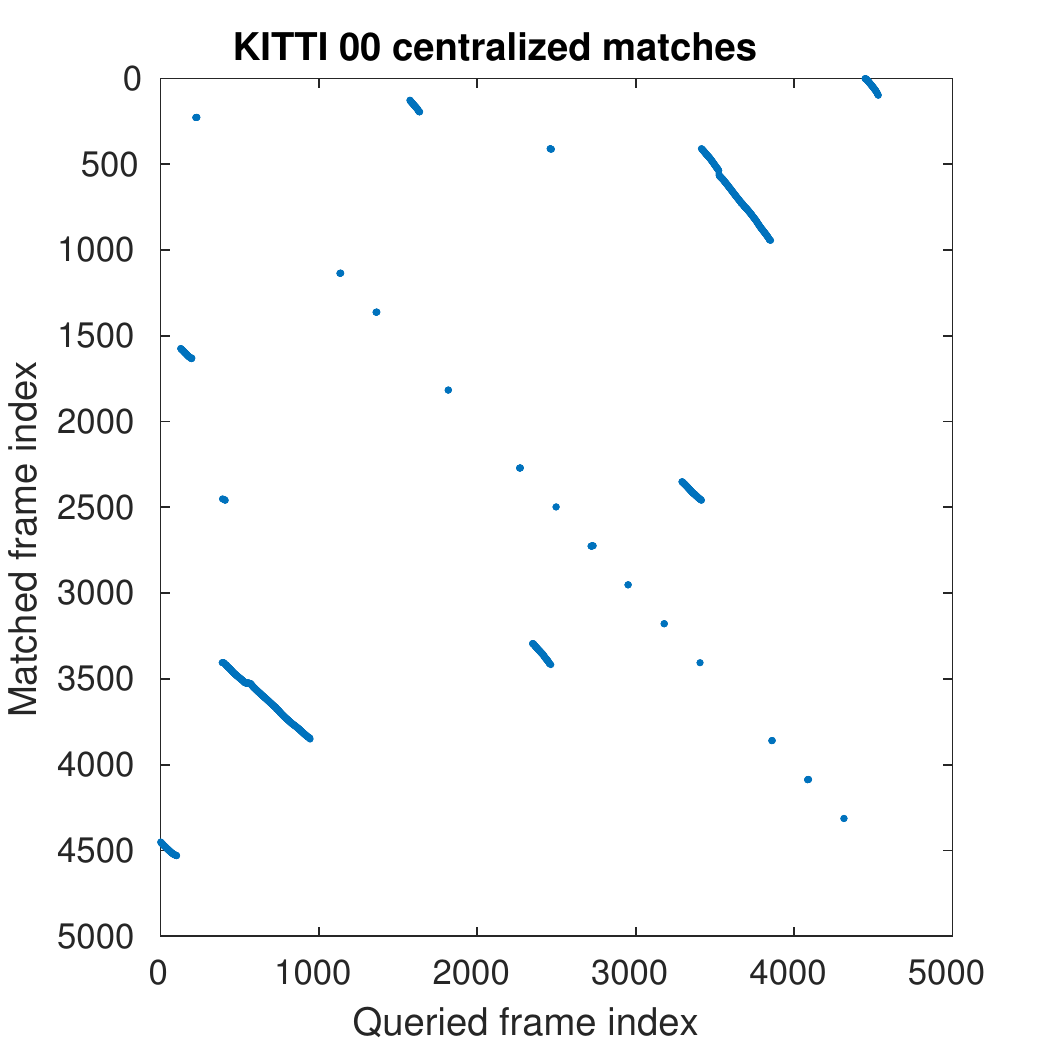}
  \caption{Confusion matrix for a centralized evaluation of the KITTI 00 dataset with 20 subtrajectories, using NetVLAD \cite{arandjelovic2016netvlad}.
  The threshold is manually selected and no geometric verification is performed.
  NetVLAD exhibits a visibly larger recall than bag-of-words method we used in \cite{cieslewski2017efficient}, Fig. 6.}
  \label{ck}
  %\vspace{-0.5cm}
\end{figure}

In this whitepaper, we instead propose to use a full-image descriptor place recognition method as a basis.
In particular, we use the recent, deep-learning based NetVLAD method \cite{arandjelovic2016netvlad} which has been shown to perform excellently even under severe appearance and viewpoint changes.
Indeed, as can be seen in the centralized evaluation of this method (Fig. \ref{ck}), its recall qualitatively looks better than the one of the BoW method we used in \cite{cieslewski2017efficient}.
NetVLAD uses a deep neural network to calculate a low-dimensional feature vector $\vec{v} \in \mathbb{R}^d$ from an input image.
Place matches can then be found by looking for the nearest vectors of other images according to the $\ell_2$ distance.

This method can now efficiently be decentralized in the following way:
\begin{enumerate}
\item Before deployment, cluster the feature vector space and assign each cluster center to a robot.
\item When querying place recognition of an image frame, calculate the feature vector and send it as query to only the robot assigned to the corresponding cluster.
\item That robot processes the query, stores it for future reference, and replies with the best matching frame identifier.
\item Send a full query to the robot that has observed that frame for geometric verification.
\end{enumerate}
Evidently, this method is leaner than the one proposed in \cite{cieslewski2017efficient}:
data is sent to only one robot, and no assumptions on the fidelity of partial responses are made.
In this preliminary work, we use k-means clustering {\color{red} cite!} as clustering method for step 1).
The clustering is trained on image data from the Oxford RobotCar dataset \cite{maddern2016year}.

\section{Experiments}

Since this is preliminary work, we use a simpler evaluation methodology than the one in \cite{cieslewski2017efficient}:
firstly, we don't actully implement the method on multiple processes as we did there.
It is evident form the method that it needs $n$ times less data exchange than if all queries were sent to all robots.
To evaluate the place recognition performance the method would have if deployed on a group of robots, we simply exclude all images that are not in the same cluster as the query from the pool of possible responses to a query.

\begin{figure}
  \centering
  \includegraphics[width=\columnwidth]{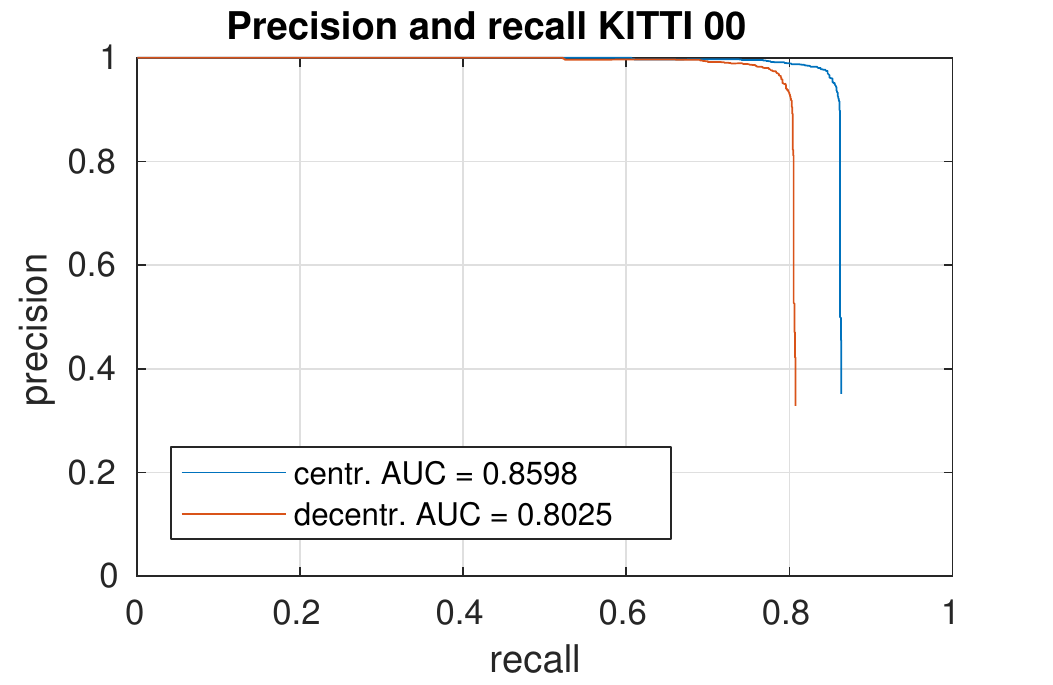}
  \caption{We evaluate the place recognition performance using the area-under-curve measure (AUC) of the precision-recall curve, since we don't apply geometric verification.
  As we can see, NetVLAD exhibits excellent precision for the most part.
  We furthermore see how the clustering of the decentralized method results in reduced recall.}
  \label{pk}
  %\vspace{-0.5cm}
\end{figure}

Secondly, we don't perform geometric verification.
To keep a fair evaluation, we evaluate precision and recall for all possible feature vector distance thresholds and consider the area under that curve (AUC) as metric for place recognition performance, see Fig. \ref{pk}.
We use then NetVLAD feature vector dimension $d = 128$ (tunable thanks to a final layer that does principal component analysis).
The method is evaluated on KITTI 00 \cite{geiger2013vision} by splitting the sequence into $n$ sub-sequences, one per robot.

\section{Results}

\begin{figure}
  \centering
  \includegraphics[width=\columnwidth]{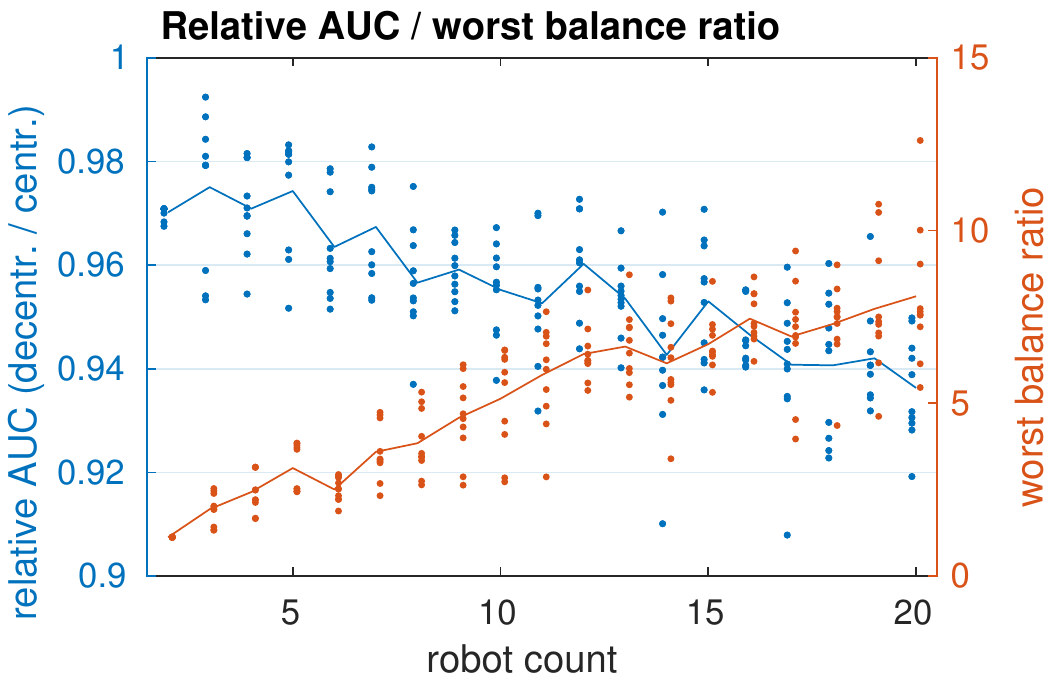}
  \caption{Relative AUC (decentralized to centralized) and worst balance ratio of our method for different robot counts.
  The worst balance ratio is the ratio of the busy-ness of the most queried robot compared to what it would be if the feature-to-cluster assignments were perfectly balanced.
  Results are averaged over 10 runs and dots indicate the results of the individual runs.}
  \label{ak}
  %\vspace{-0.5cm}
\end{figure}

Fig. \ref{ak} shows relative AUC (decentralized to centralized) of the method when applied to groups of $n \in [2, 20]$ robots.
Recall suffers if the true match of a query is not in the same cluster as the query.
It would seem that the performance of the decentralized NetVLAD method is only marginally better than the performance of the decentralized BoW method (see Fig. 7 in \cite{cieslewski2017efficient}).
Consider however that as qualitatively seen in Fig. \ref{ck}, NetVLAD already has a higher recall than BoW in the first place.
Furthermore, the method uses far less bandwidth for its distributed query than the BoW method.
Recall from Table II in \cite{cieslewski2017efficient} that its distributed query size is $16$ kilobytes plus overhead from sending the query to $n$ robots.
This method, when using single precision, only needs $d \times 4$ bytes per query, so $512$ bytes with $d = 128$, plus overhead from only sending to one robot.

However, as it turns out, this method at its current state is very bad at balancing the computational and network load among the robots.
In Fig. \ref{ak} we furthermore report the {\it worst balance ratio}, a measure for how much more queries the busiest robot receives compared to how much it would receive if the queries were perfectly balanced.
As we can see, the busiest robot handles up to half of all queries!
This can be traced back to bad clustering.

\begin{figure}
  \centering
  \includegraphics[width=\columnwidth]{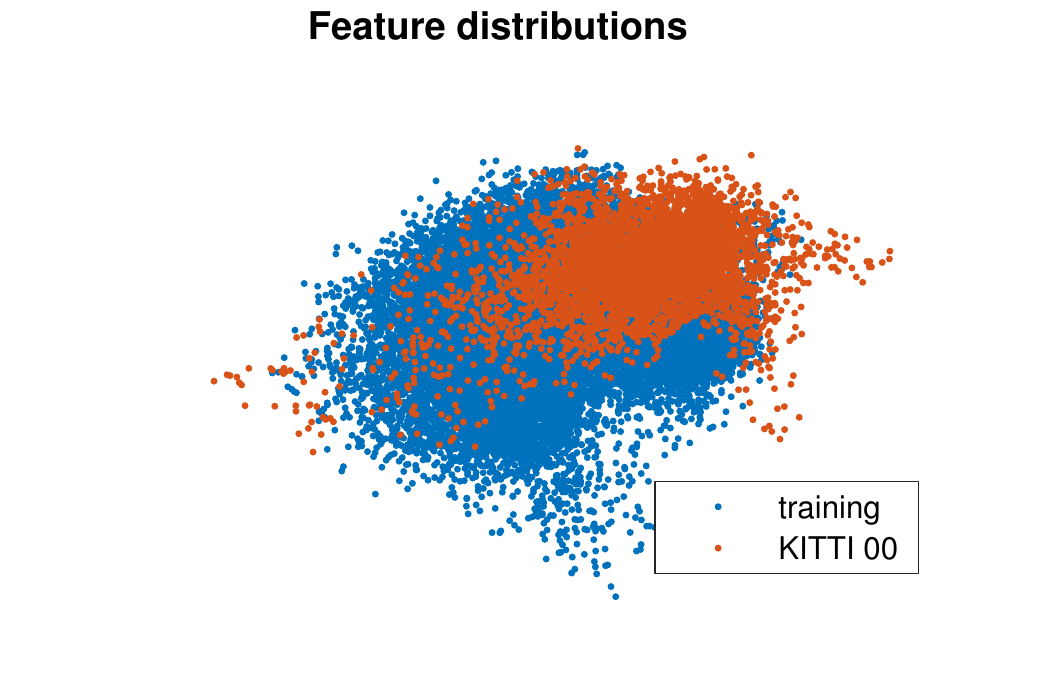}
  \caption{The bad load balancing during evaluation can be explained by the difference in distribution of the image features.
  This is the distribution of the first two dimensions for the training and testing data.
  Training data should be more general / cover more environments than the deployment data, which can only come from a very specific kind of environment.
  Hence, this is a problem that is inherent in the method.}\label{dk}
  %\vspace{-0.5cm}
\end{figure}

Why is the clustering bad?
This can happen when the distribution of feature vectors in the deployment dataset is not the same as the distribution of vectors in the cluster training dataset.
As it turns out, the features in KITTI 00 only occupy a subspace of what the features in the RobotCar dataset occupy, see Fig. \ref{dk}.
This is however not a problem of the training but a problem that is implicit in the method:
The training set should be as general as possible (i.e. covering as many different environments as possible), while it is perfectly possible for the feature vectors encountered during deployment to stem only from a very specific type of environment.

In future work we will try to overcome the balancing issue, and try to see what parts of the pipeline can be improved to yield better performance.

\section*{Acknowledgement}

We would like to thank Antonio Loquercio for the code reviews and helpful feedback.

%%%%%%%%%%%%%%%%%%%%%%%%%%%%%%%%%%%%%%%%%%%%%%%%%%%%%%%%%%%%%%%%%%%%%%%%%%%%%%%%

\bibliographystyle{IEEEtran}
\bibliography{references}

\end{document}